\documentclass{article}

 \usepackage[preprint, nonatbib]{neurips_2023}

\usepackage[utf8]{inputenc} 
\usepackage[T1]{fontenc}    
\usepackage{hyperref}       
\usepackage{url}            
\usepackage{booktabs}       
\usepackage{amsfonts}       
\usepackage{nicefrac}       
\usepackage{microtype}      
\usepackage{xcolor}         
\usepackage{amsmath}
\usepackage{multirow}
\usepackage{comment}
\usepackage{wrapfig}
\usepackage{graphicx}

\title{UniCat: Crafting a Stronger Fusion Baseline for Multimodal Re-Identification}

\author{
  Jennifer Crawford \\
  Modern Intelligence\\
  \texttt{jenni@modernintelligence.ai}\\
  \And
  Haoli Yin \\
  Modern Intelligence\\
  \texttt{haoli@modernintelligence.ai}\\
  \And
  Luke McDermott \\
  Modern Intelligence\\
  \texttt{luke@modernintelligence.ai}\\
  \And
  Daniel Cummings \\
  Modern Intelligence\\
  \texttt{daniel@modernintelligence.ai}
}

\begin{document}

\maketitle

\begin{abstract}
Multimodal Re-Identification (ReID) is a popular retrieval task that aims to re-identify objects across diverse data streams, prompting many researchers to integrate multiple modalities into a unified representation. While such fusion promises a holistic view, our investigations shed light on potential pitfalls. We uncover that prevailing late-fusion techniques often produce suboptimal latent representations when compared to methods that train modalities in isolation. We argue that this effect is largely due to the inadvertent relaxation of the training objectives on individual modalities when using fusion, what others have termed modality laziness. We present a nuanced point-of-view that this relaxation can lead to certain modalities failing to fully harness available task-relevant information, and yet, offers a protective veil to noisy modalities, preventing them from overfitting to task-irrelevant data. Our findings also show that unimodal concatenation (UniCat) and other late-fusion ensembling of unimodal backbones, when paired with best-known training techniques, exceed the current state-of-the-art performance across several multimodal ReID benchmarks. By unveiling the double-edged sword of "modality laziness", we motivate future research in balancing local modality strengths with global representations. 
\end{abstract}

\section{Introduction}
Multimodal representation learning has rapidly grown in popularity as deep learning researchers are continually innovating new ways to combine different channels of information simultaneously. Common tasks in this field revolve around using text and images for input/output \cite{akkus2023multimodal} or even using image-paired data as a basis for learning a joint embedding across a wide variety of modalities (e.g., text, audio, IMU, depth) \cite{girdhar2023imagebind}. However, an often overlooked subfield is in the area of multimodal re-identification (ReID), where the objective is to match instances of the same object across different sensor sources and type (e.g., visible spectrum, infrared). In many cases, ReID entails learning object representations from multiple views such as in security or retail applications, where one might want to identify if an object appearing in one camera's field-of-view is the same object appearing in another camera later in time. In this setting, each modality can have its strengths and limitations. For instance, although a regular RGB image has color information, it will be affected by illumination changes (e.g., daytime vs. nighttime), whereas an infrared image might not.

Despite their ambitious goals, many recent fusion methods encounter limitations. A notable challenge is \textit{modality laziness} as described by Du et al. \cite{du2023unimodal}. In conventional late fusion techniques that employ shared loss functions between modalities—such as vanilla late fusion through concatenating or averaging representations, where the cross-entropy (CE) loss is determined after fusion \cite{Ye2020ReIDSurvey}—there is potential for individual modalities to contribute less to the overall training classification accuracy. This concern is especially evident in downstream retrieval tasks like ReID. Based on findings by Peng et al. \cite{peng2022balanced}, one could infer that when modalities are trained individually, the representation of a single modality will have a direct influence on the loss function since it's the sole contributor in that setting.

Guided by these challenges, this study zeroes in on evaluating fusion techniques, especially within the ReID context. While our lens is sharply focused on ReID, the derived insights hold broader relevance, extending to a vast spectrum of multimodal tasks that utilize late fusion techniques. Our main contributions begin with creating a comprehensive evaluation of late fusion techniques in the current multimodal ReID landscape, spotlighting their inherent limitations from a \textit{modality laziness} perspective. Next, we provide empirical evidence establishing late-fusion unimodal ensembled models as a new strong baseline, outperforming existing state-of-the-art methods in multimodal ReID benchmarks. Finally, we discuss plausible explanations for this surprising performance behavior in the context of unimodal model ablations and future research directions for better understanding multimodal representation learning and model design.

\section{Preliminaries and Method}

Multimodal representation learning focuses on the efficient and effective compression of high-dimensional data, especially for downstream tasks like ReID. For this purpose, convolutional neural networks (e.g., ResNet-50) and Vision Transformers (ViTs) are typically employed as visual encoders \cite{mehta2020}. Within this context, \textit{late fusion} stands out as the predominant method for integrating multiple data modalities. This approach is favored over the less-common \textit{early fusion}, which integrates multiple data modalities before inputting them into a single model. Notably, early fusion can be sensitive to pre-processing and lacks scalability \cite{2017Ramachandram}. The late fusion approach first encodes each modality separately before combining their outputs, usually via averaging or concatenation mechanisms. While late fusion does not allow for cross-modal interactions, it is one of the most widespread fusion strategies due to the intuition that the approach is able to leverage already existing and carefully optimized architectures for the distinct modalities.

In object ReID, the task is to match objects across different data captures, typically from disparate views or time instances. Multimodal ReID datasets are given by $n$ object samples, where each sample consists of $M$ coincidental images (corresponding to $M$ modalities) and an identifying label (ID): for each sample we have $[(x_1, x_2, ..., x_M), y]$ where $x_i \in \mathbb{R}^{C \times H \times W}$ is the $i$-th modality image of the object and  $y\in \mathcal{D}$ denotes its ID. We consider \textit{multiple views} of an object to be different samples that share the same object ID. Of note, the IDs we see during train-time do not overlap with those seen at inference. As such, object ReID is often solved not by using the direct output of a classifier, but by bringing different views of the same object together in the latent space and push different objects away from each other. This is commonly done by combining a cross-entropy loss with contrastive learning in the form of triplet loss:
\begin{equation}
\mathcal{L}(z) = \mathcal{L}_{\text{tri}}(z) + \lambda \mathcal{L}_{\text{CE}}(\tilde{y}, y)
\label{eq:reidloss}
\end{equation}
where $\mathcal{L}_{\text{tri}}$ is a soft-margin triplet loss (further defined in Appendix B) that uses the object embedding $z$ and $\mathcal{L}_{\text{CE}}$ denotes a cross-entropy loss that uses the output of a linear classifier, $\tilde{y}$. Here, $\lambda$ acts as a balancing coefficient.

To establish a strong baseline for future multimodal ReID research and investigate the impacts of fusion, we study three separate vanilla late-fusion strategies: fusion by concatenation (\textit{Fusion-concat}), fusion by averaging (\textit{Fusion-avg}), and post-training fusion by unimodal concatenation (\textit{UniCat}). 

More specifically, for each modality we train an unshared backbone, $f_i$ that produces a modality-specific embedding, $z_i$.
These modality-specific embeddings can then be combined into a multimodal representation: $z_{\text{fuse}} = \Theta(z_1, z_2, ... z_M)$ where $\Theta$ is the chosen fusion operator. The difference among the training strategies lies in whether global or local loss functions are used. In \textit{Fusion-avg} or \textit{Fusion-concat}, we use a global loss function that operates on the fused multimodal representation, $z_{\text{fuse}}$: $\mathcal{L}_{\text{fusion}}=\mathcal{L}(z_{\text{fuse}}, \tilde{y}_{\text{fuse}})$. In this case, the back-propagated gradients for each modality are entangled as described in \cite{peng2022balanced}. Conversely, in \textit{UniCat} we train each modality encoder independently with their own local loss function: $\mathcal{L}_{\text{post}}=\sum_i^M \mathcal{L}(z_i, \tilde{y}_i)$ where $i$ is summed over all modalities. Here, note that each modality's loss is fully disentangled from the others. In this case, fusion occurs only at inference when we concatenate the modality-specific embeddings into the final multimodal representation, $z_{\text{fuse}}$. For additional details on the training strategies used, please see Appendix B.


\section{Results}

\begin{wrapfigure}{r}{0.35\textwidth}
  \centering
  \includegraphics[width=1.7in]{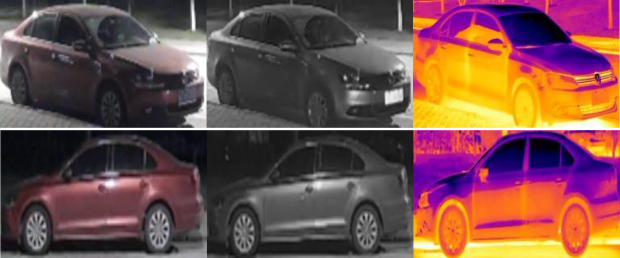}
  \caption{RGB, NIR, TIR \\\hspace{\textwidth} images from RGBNT100 \cite{Li-2020a}.}
  \label{fig:100_examples}
\end{wrapfigure}

We benchmark the described late-fusion strategies on three multimodal ReID datasets. RGBNT100 \cite{Li-2020a} is a multimodal vehicle dataset consisting of coincidental visible (RGB), near-infrared (NIR), and thermal-infrared (TIR) images for 100 vehicles from multiple views as shown in Figure \ref{fig:100_examples}. RGBN300 is an extension of RGBNT100 in which (RGB, NIR) pairs are provided for 300 vehicles. Lastly, RGBNT201 \cite{Zheng2021} is a multimodal person dataset that provides (RGB, NIR, TIR) image triples for 201 different individuals. The key benchmarks, mAP and Rank-1, are informed by the cumulative match curve (CMC) \cite{zheng2015scalable}. Notably, mAP offers an overarching performance measure.

We compare our late-fusion baseline strategies with the previous state-of-the-art multimodal ReID models (described in the Appendix A). In particular, we test two of the most widely-used backbones found in other ReID models, ResNet-50 and ViT-B.

In Table \ref{tab:MMresults} we observe that RGBNT100/RGBN300 with \textit{UniCat} and RGBNT201 with \textit{Fusion-concat} do strikingly better than the other best-known fusion approaches, achieving state-of-the-art performance. We observe backbone choice to have a more dramatic impact on RGBNT201 than RBGNT100/RGBN300. Since RGBNT201 contains 2x fewer datapoints than RGBNT100, data-hungry ViTs may struggle to learn with lack of inductive biases compared to CNNs \cite{zhu2023understanding}. Overall, we clearly observe fusion strategy as a leading factor on representation quality. 

\begin{table}[htb]
\small
\begin{center}
\noindent\makebox[\textwidth]{%
\begin{tabular}{c c  c  c  c  c c c } 
\hline
\multirow[c]{2}{*}{Model} & \multirow[c]{2}{*}{Backbone} & \multicolumn{2}{c}{RGBNT100} & \multicolumn{2}{c}{RGBN300} & \multicolumn{2}{c}{RGBNT201}\\
\cline{3-8}
 & & mAP & Rank-1 & mAP & Rank-1 & mAP & Rank-1\\
\hline
PFNET \cite{Zheng2021} & ResNet-50 & 68.1$^\dag$ & 94.1$^\dag$ & - & - & 31.8 & 54.6 \\
IEEE \cite{IEEE} & ResNet-50 & 61.3$^\dag$ & 87.8$^\dag$ & - & - & 46.42 & 47.13 \\
HAMNet \cite{Li-2020a} & ResNet-50 & 64.1 & 84.7 & 61.9 & 84.0 & 27.68$^*$ & 26.32$^*$ \\
CCNet \cite{CCNET} & ResNet-50 & 77.2 & 96.3  & - & -  & - & - \\
PHT \cite{PHT} & ViT-B & 79.9 & 92.7 & 79.3 & \textbf{93.7} & - & - \\
\hline
Fusion-avg & ResNet-50 & 75.3 \(\pm\)0.6 & 92.2 \(\pm\)0.9 & 74.9 \(\pm\)0.5 & 91.2 \(\pm\)0.6 & 57.6 $\pm$2.9 & 61.0 $\pm$3.0 \\
Fusion-concat & ResNet-50 & 75.1 \(\pm\)0.9 & 92.7 \(\pm\)0.6 & 74.7 \(\pm\)0.4 & 91.3 \(\pm\)0.2 & \textbf{63.0 $\pm$1.9} & \textbf{66.7 $\pm$3.1}  \\
UniCat & ResNet-50 & 79.4 \(\pm\)0.5 & 94.9 \(\pm\)0.8 & 76.7 \(\pm\)0.4 & 92.3 \(\pm\)0.4 & 38.1 $\pm$1.3 & 33.6 $\pm$1.7 \\
Fusion-avg & ViT-B & 76.1 \(\pm\)0.3 & 90.1 \(\pm\)0.9 & 78.1 \(\pm\)0.1 & 91.4 \(\pm\)0.4 & 56.1 \(\pm\)2.2 & 54.9 \(\pm\)1.8 \\
Fusion-concat & ViT-B & 75.9 \(\pm\)0.7 & 92.3 \(\pm\)0.4  & 77.4 \(\pm\)0.3 & 91.8 \(\pm\)0.3 & 61.7 \(\pm\)2.1 & 60.6 \(\pm\)1.2 \\
UniCat & ViT-B & \textbf{81.3 \(\pm\)0.9} & \textbf{97.5 \(\pm\)0.3}  & \textbf{80.2 \(\pm\)0.2} & 92.9 \(\pm\)0.4 & 57.6 \(\pm\)2.0 & 58.0 \(\pm\)1.8 \\
\hline
\end{tabular}}
\end{center}
\caption{Comparison between current state-of-the-art multimodal ReID models with our strong multimodal baselines. Metrics for our models were calcluated by averaging performance across four random seeds. $\dag$, $*$ indicates the results were taken from \cite{CCNET} and \cite{IEEE}, respectively.}
\label{tab:MMresults}
\end{table}

To gain intuition for the dynamics at play, we further compare all of our modality backbones on the task of unimodal ReID. That is, we use our trained modality encoders from Table \ref{tab:MMresults} and evaluate unimodal ReID performance on the test sets. Because the multimodal embeddings in all fusion cases are simply poolings of unimodal embeddings, multimodal ReID performance here is closely tied to the corresponding unimodal ReID performance of the constituent modalities. As shown in Table \ref{tab:single100}, for RGBNT100 we observe that all unimodal ReID performance has been negatively affected by training with fusion. Surprisingly, we also observe this to be the case in RGBNT201, with the exception of one modality - NIR (see Table \ref{tab:single201}). We can consider this modality a weak-link since when trained in isolation (\textit{UniCat}) its performance is notably lower than RGB/TIR. As multimodal ReID is most affected by the weakest-link, we deduce that the performance gains of NIR in the case of \textit{Fusion-concat} outweighed the performance drops in RGB/TIR. 

We suspect these effects are largely due to the impact and nuances of \textit{modality laziness}\footnote{For insight into the fundamental nature by which modality laziness operates, see Appendix C for an experiment dealing with single-modality ensembles.}. To confirm that this underlies what we observe, as opposed to overfitting as speculated by \cite{gblend}, we evaluate the embeddings of the RGBNT100 training set and find that they achieve a lower unimodal ReID performance in the cases of training with fusion (Appendix D). The results shown by RGBNT100/RGBN300 demonstrate the negative effects that modality laziness can have - leading to  underutilized unimodal features. Conversely, for datasets such as RGBNT201 where noisy, less-reliable modalities exist, modality laziness may offer a training relief by helping to prevent those modalities from resorting to high-degrees of overfitting. This latter notion has also been hypothesized by Liu et al. \cite{noisy}.

\begin{table}[tb]
\small
\begin{center}
\noindent\makebox[\textwidth]{%
\begin{tabular}{ c c  c  c  c  c  c  c } 
\hline
\multirow[c]{2}{*}{Model} & \multirow[c]{2}{*}{Backbone}&  \multicolumn{2}{c}{RGB} & \multicolumn{2}{c}{NIR} & \multicolumn{2}{c}{TIR}\\
\cline{3-8}
 & & mAP & Rank-1 & mAP & Rank-1 & mAP & Rank-1 \\
 \hline
 Fusion-avg & ResNet-50 & 48.6 $\pm$1.4 & 71.9 $\pm$0.9 & 40.4 $\pm$0.5 & 62.3 $\pm$1.3 & 41.1 $\pm$1.1 & 71.0 $\pm$2.3 \\
 Fusion-concat & ResNet-50 & 44.7 $\pm$1.2 & 63.9 $\pm$2.8 & 27.3 $\pm$1.0 &  43.5 $\pm$1.1 & 36.9 $\pm$1.5 & 62.9 $\pm$1.3 \\
 UniCat & ResNet-50 & 55.0 $\pm$0.8 & 77.2 $\pm$0.9 & 46.0 $\pm$1.1 & 67.3 $\pm$1.0 & \textbf{52.5 $\pm$0.4} & \textbf{82.5 $\pm$1.3} \\
 Fusion-avg & ViT-B & 51.7 \(\pm\)0.8 & 73.5 \(\pm\)0.9 & 41.8 \(\pm\)0.6 & 59.4 \(\pm\)2.7 & 31.0 \(\pm\)0.9 & 53.6 \(\pm\)2.8 \\
 Fusion-concat & ViT-B & 50.0 \(\pm\)4.1 & 70.8 \(\pm\)5.6 & 38.9 \(\pm\)2.5 & 60.7 \(\pm\)1.5 & 34.1 \(\pm\)1.6 & 61.6 \(\pm\)3.0\\
 UniCat & ViT-B & \textbf{58.4 \(\pm\)0.83} & \textbf{80.2 \(\pm\)1.2} & \textbf{51.4 \(\pm\)0.8} & \textbf{71.5 \(\pm\)1.5} & 46.1 \(\pm\)2.5 & 76.0 \(\pm\)3.2\\
\hline
\end{tabular}}
\end{center}
\caption{RGBNT100 \cite{Li-2020a} Evaluation of unimodal ReID using the embeddings learned from our multimodal late-fusion ReID baselines.}
\label{tab:single100}
\end{table}

\begin{table}[tb]
\begin{center}
\small
\noindent\makebox[\textwidth]{%
\begin{tabular}{ c  c c  c  c  c  c  c } 
\hline
\multirow[c]{2}{*}{Model} & \multirow[c]{2}{*}{Backbone}& \multicolumn{2}{c}{RGB} & \multicolumn{2}{c}{NIR} & \multicolumn{2}{c}{TIR}\\
\cline{3-8}
 & & mAP & Rank-1 & mAP & Rank-1 & mAP & Rank-1 \\
 \hline
 Fusion-avg & ResNet-50 & 21.4 $\pm$1.8 & 17.5 $\pm$3.0 & 14.8 $\pm$1.0 & 11.5 $\pm$0.6 & 33.0 $\pm$1.9 & 33.6 $\pm$3.5\\
 Fusion-concat & ResNet-50 & 16.7 $\pm$2.3 & 14.2 $\pm$2.8 & 18.9 $\pm$2.1 & 16.6 $\pm$2.8 & 28.5 $\pm$0.6 & 30.0 $\pm$1.5\\
 UniCat & ResNet-50 & 23.2 $\pm$2.0 & 18.34 $\pm$2.5 & 11.5 $\pm$0.6 & 7.4 $\pm$0.3 & 36.0 $\pm$0.3 & 36.9 $\pm$1.2\\
 Fusion-avg & ViT-B & 28.1 $\pm$1.8 & 25.2 $\pm$1.6 & 20.5 $\pm$2.2 & 16.5 $\pm$3.0 & 27.6 $\pm$0.8 & 26.2 $\pm$1.1\\
 Fusion-concat & ViT-B & 28.2 $\pm$2.7 & 25.8 $\pm$2.8 & \textbf{24.7 $\pm$1.5} & \textbf{23.7 $\pm$0.9} & 30.0 $\pm$0.8 & 29.0 $\pm$1.5\\
 UniCat & ViT-B & \textbf{31.9 $\pm$0.8} & \textbf{29.6 $\pm$1.2} & 20.9 $\pm$1.1 & 16.8 $\pm$1.5 & \textbf{37.5 $\pm$1.3} & \textbf{36.2 $\pm$1.6}\\
\hline
\end{tabular}}
\end{center}
\caption{RGBNT201 \cite{Zheng2021} Evaluation of unimodal ReID using the embeddings learned from our multimodal late-fusion ReID baselines.}
\label{tab:single201}
\end{table}
\section{Discussion}
Our results indicate that when different modalities are jointly optimized (fusion training), they may not capture all available task-relevant information, leading to potential underrepresentation. This is especially pronounced in datasets like RGBNT100/RGBN300. But there's a silver lining: while certain modalities suffer, noisy ones may benefit since joint training seems to act as a form of regularization for such modalities, preventing overfitting to task-irrelevant data. Thus, a delicate equilibrium emerges in multimodal ReID learning. We find that the \textit{UniCat} approach using a ViT-B backbone outperforms in both RGBNT100 and RGBN300. Here, each modality—RGB, NIR, and TIR—benefits from the pressure of being trained in isolation. Challenging views like those in \cite{gblend} which focus on over-parameterization, our results suggest that joint optimization's nuances may limit modalities from capturing complete task-relevant information, therein lying the negative effects of modality laziness. Conversely, in certain scenarios, such as with the RGBNT201 dataset, the utility of modality laziness emerges. Specifically, the \textit{Fusion-concat} method bolsters the performance of NIR—a notably weaker modality. While NIR struggles when optimized individually, joint optimization uplifts its performance. This could be attributed to modality laziness acting as a shield against the modality's inherent noise, resonating with the observations made by \cite{noisy}. This not only highlights the protective role of modality laziness but also underscores its potential as a regularization mechanism, especially for modalities with less task-relevant information. Previous solutions to modality laziness (\cite{du2023unimodal}, \cite{peng2022balanced}) do not consider the full scope of the phenomenon.

Our findings show that while our simple baseline multimodal ReID models outperform existing state-of-the-art, much research is still needed to inform more holistic, nuanced multimodal frameworks. We encourage researchers in this field to validate simple post-training fusion baselines like \textit{UniCat} first in their setting, before pursuing more complex fusion strategies. Lastly, frameworks that are both model parameter efficient (e.g., a shared backbone encoder) and data efficient are essential for future architecture scalability.

\newpage

\small
\bibliographystyle{plain}
\bibliography{bibliography.bib}
\normalsize

\newpage
\appendix

\section{Related Work}
In current multimodal ReID literature, fusion approaches can be broadly categorized into late and middle fusion. Late fusion entails processing each modality independently and merging the representations in the final stages. Middle fusion, a subtype of late fusion, involves interchanging features between modalities to create an intricately fused representation upon which subsequent tasks are executed. This nuanced approach in middle fusion allows for a more integrated analysis of modalities compared to traditional late fusion.

\subsection{Late Fusion Approaches: }
\textbf{HAMNet} \cite{Li-2020a}. The paper introduces HAMNet, a multi-stream convolutional network (i.e. separate ResNet50 for each modality) tailored for multimodal vehicle ReID. It employs multiple backbones to independently process each modality; however, their method joins the features from each modality via a modified pooling operation and uses a global loss function on the joint representation, which suffers from modality laziness per our previous discussion. 

\textbf{CCNet} \cite{CCNET}. CCNet, as described in the paper, is a multimodal vehicle ReID model that employs a multi-stream architecture, with each branch dedicated to processing individual modalities. This method mirrors the approach of UniCat through its use of modality-specific cross entropies. However, CCNet differs by incorporating global contrastive losses, which is suboptimal.

\textbf{PHT} \cite{PHT}. The paper introduces the Progressively Hybrid Transformer (PHT) for multimodal vehicle ReID. This method employs individual transformer branches for each modality and integrates them at a singular depth, while utilizing the Modal-specific Controller (MC) and Modal Information Embedding (MIE) for modality adjustments. This design embodies late fusion, given the independent modality processing and a singular fusion point using averaging. Notably, this was the previous state-of-the-art method and is the first transformer-based method to tackle the multimodal ReID problem. However, late fusion with averaging prevents each modality from extracting the maximum amount of task-relevant information per modality laziness. 

\subsection{Middle Fusion Approaches: }
\textbf{PFNET} \cite{Zheng2021}. The Progressive Fusion Network (PFNet) for multimodal person ReID follows a middle fusion approach, initially processing modalities separately but later integrating them through intermediate and global stages. Despite its middle fusion characteristics, the architecture predominantly relies on simple concatenation of latent factors, which may not be optimally aligned for the desired tasks. This suggests a potential need for a more constrained inter-modality interaction and balance in learning effective representations. 

\textbf{IEEE} \cite{IEEE}. The Interact, Embed and EnlargE (IEEE) method for multimodal person ReID employs independent modality processing, followed by cross-modal interactions and feature enhancements. This middle fusion strategy is evident as it interchanges and integrates modality-specific features at intermediate stages. However, its reliance on basic feature summation and late-stage concatenation suggests further investigation into how to best balance the trade-offs to alleviate modality laziness. 


\section{Training Details}

For training, we combine triplet loss \( \mathcal{L}_{\text{tri}} \) \cite{TransReID} that tries to increase class separability of the embeddings $z$ with a cross-entropy loss \( \mathcal{L_{\text{CE}}} \) that seeks to put the embeddings into class-specific subspaces \cite{BoT} (see eqn \ref{eq:reidloss}). Specifically, we use a modified soft-margin triplet loss that includes hard-mining \cite{xuan2021hard} to select the most challenging positive and negative samples ($p$ and $n$) for a given anchor ($a$):
\begin{equation}
\mathcal{L}_{\text{tri}} = \min_n\max_p \log\left(1 + \exp\left(||a-p||_2 - ||a-n||_2 + \alpha\right)\right)
\label{eq:triplet}
\end{equation}
For the cross-entropy (CE) loss, class-wise output scores are found via a linear classifier, and then compared against the ground truth. When combining these losses in eqn \ref{eq:reidloss} we set the balancing coefficient \( \lambda \) to 1.

Images are resized to 128x256 for RGBNT100/RGBN300 and 256x128 for RGBNT201. During training, images are augmented with random horizontal flipping, padding, random cropping and random erasing \cite{TransReID}. We use many of the training strategies suggested by \cite{BoT}, including a linear warmup strategy for the learning rate and using BNNeck to normalize features before CE loss. For each fusion strategy, we perform grid-search hyperparameter optimization across three batch sizes ([64, 128, 256]) and three learning rates ([.008, .016, .032]). We train for 120 and 200 epochs for ViT-B and ResNet-50 backbones, respectively, used SGD with momentum 0.9 and a cosine learning rate decay. All backbones are instantiated with Imagenet \cite{imagenet} pre-trained weights while the linear classifier and bottleneck layers are instantiated via \cite{kaiming}. Similar to \cite{BoT}, we choose cosine distance as our similarity metric for ReID.

\section{Effects of Fusion Strategies in Ensemble Learning}

\begin{table}[htb]
\small
\begin{center}
\begin{tabular}{ c  c  c  c  c  c  c  c  c }
\hline
\multirow[c]{3}{*}{Model} & \multicolumn{6}{c}{RGBNT100} & \multicolumn{2}{c}{\multirow[c]{2}{*}{Market1501}}\\
\cline{2-7}
 & \multicolumn{2}{c}{RGB} & \multicolumn{2}{c}{NIR} & \multicolumn{2}{c}{TIR} & \multicolumn{2}{c}{} \\
 \cline{2-9}
 & mAP & Rank-1 & mAP & Rank-1 & mAP & Rank-1 & mAP & Rank-1\\
 \hline
Fusion avg & 57.7 & 79.1 & 48.6 & 67.7 & 48.6 & 76.9 & 87.6 & 94.9\\
Fusion-concat & 58.4 & 81.0 &  48.9 & 70.0 & 47.4 & 80.8 & 87.9 & 94.8\\
UniCat & \textbf{60.4} & \textbf{84.4} & \textbf{50.7} & \textbf{72.9} & \textbf{49.7} & \textbf{81.1} & \textbf{89.1} & \textbf{95.4}\\
\hline
\end{tabular}
\end{center}
\caption{Comparison of late-fusion implementations for a two-backbone (ViT-B) unimodal ensemble learner.}
\label{tab:ensemble}
\end{table}

Peng et al. \cite{peng2022balanced} suggests that modality laziness occurs when some modalities are more dominant than others. However, we suspect that this phenomenon can occur even when modalities have comparable, or even equal, predictive strengths. To demonstrate this idea, we look to ensembles. Specifically, we train an ensemble of unimodal ReID models on a single modality. When training ensembles, it is common practice to train each ensemble member individually and fuse their predictions only at inference (i.e. bagging). We compare this standard strategy of training each ensemble member independently (\textit{UniCat}) with training them jointly by having a global loss use their fused output embedding (\textit{Fusion-avg/concat}). 

For this experiment, we investigate each modality within RGBNT100 as well as the popular unimodal person ReID dataset, Market1501 \cite{market1501}. We choose ViT-B as our model backbone and use ImageNet pre-trained weights. The classifiers and bottlenecks are randomly instantiated via \cite{kaiming}, so that we may fully expect the backbones to have unique final solutions.

As shown in Table \ref{tab:ensemble}, there is a boost in unimodal RGBNT100 ReID performance for all ensemble models when compared to Table \ref{tab:single100}, which is expected as ensembling helps to diminish the impacts of overfitting. However, what is notable is that in all unimodal tasks, training the ensemble using a global loss (\textit{Fusion-avg/concat}) resulted in a lower performance than would otherwise have been obtained if each member were trained independently (\textit{UniCat}). This underscores the fundamental nature by which modality laziness operates and how the phenomenon should be considered even when training with balanced modalities. 

\section{RGBNT100 Training Performance}

\begin{figure}[h]
  \centering
  \includegraphics[width=5.5in]{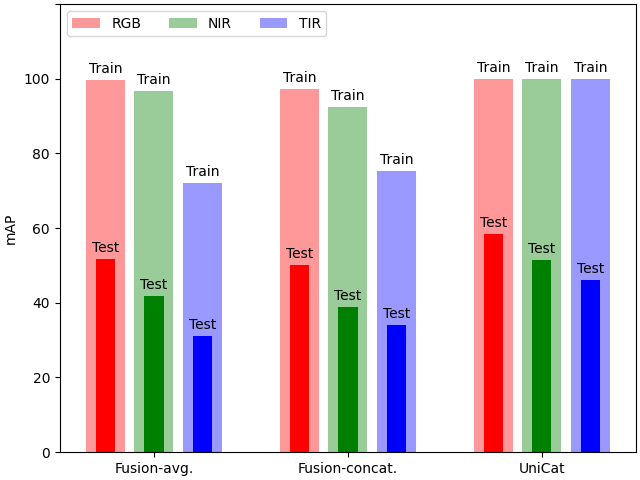}
  \caption{Comparison of train and test unimodal ReID performance for our late-fusion mulitmodal baselines on RGBNT100.}
  \label{fig:train}
\end{figure}

To further distinguish whether modality laziness or general overfitting is at play in the performance drop observed when training RGBNT100 with fusion, we evaluate the unimodal ReID performance (given by mAP) of the train set for RGBNT100 (see Figure \ref{fig:train}). Mirroring the test behavior, we see that the unimodal mAP for the train set was inhibited for \textit{Fusion-avg} and \textit{Fusion-concat} compared to \textit{UniCat}. Thus, contrary to the hypothesis of \cite{gblend}, the performance decline that we observed when training with fusion (Table \ref{tab:MMresults}) can be primarily linked to modality laziness, and it is actually \textit{UniCat} that appears to be more at risk for overfitting. This also helps to demonstrate how the impacts of modality laziness may at times be beneficial for weaker, less-informative modalities that are especially overfitting-prone, as we observed in RGBNT201 (Tables \ref{tab:MMresults} and \ref{tab:single201}).

\end{document}